\documentclass[letterpaper]{article}
\usepackage{natbib,alifeconf}

\newcommand\blfootnote[1]{%
  \begingroup
  \renewcommand\thefootnote{}\footnote{#1}%
  \addtocounter{footnote}{-1}%
  \endgroup
}

\usepackage{hyperref}
\usepackage{authblk}
\usepackage{titlepic}
\usepackage{caption}
\usepackage{float}
\usepackage[T1]{fontenc} % ??? QQQ -- "<"
\usepackage{xfrac}

\graphicspath{ {images/} }

\newcommand{\code}[1]{\textbf{\small \texttt{#1}}}

\begin{document}

\title{EvoFlock: evolved inverse design of multi-agent motion}

\author{Craig Reynolds\authorcr
    unaffiliated researcher\authorcr 
    cwr@red3d.com}

%%%%%%%%%%%%%%%%%%%%%%%%%%%%%%%%%%%%%%%%%%%%%%%%%%%%%%%%%%%%

\captionsetup{hypcap=false}

\titlepic{\includegraphics[width=\textwidth]{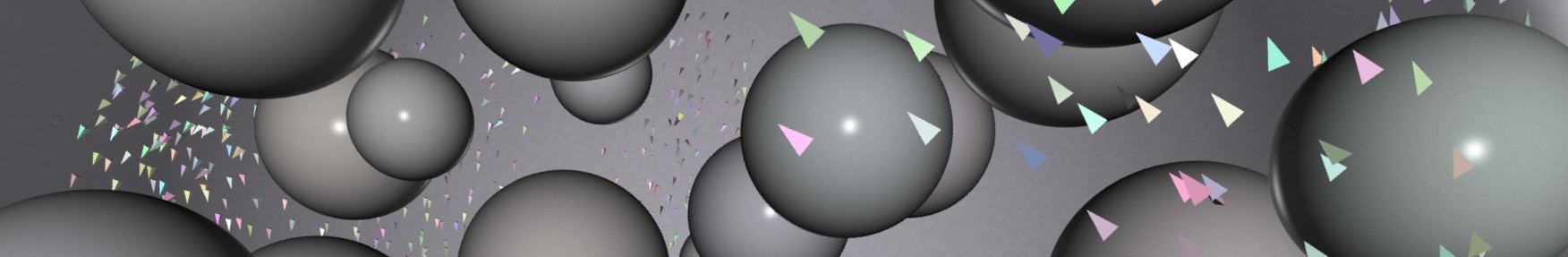}
\captionof{figure}{1000 boids flocking in a space cluttered with obstacles. Behavioral parameters of the boids are determined by inverse design, using multi-objective evolutionary optimization. See this simulation running in \href{https://youtu.be/XkwJXPJwxIc}{Video 1}.} 
\label{fig:boid_flock}}

% Remove today's date being inserted after the title/author information.
\date{}

%% Lay out the single column top matter defined above.
\maketitle

%%%%%%%%%%%%%%%%%%%%%%%%%%%%%%%%%%%%%%%%%%%%%%%%%%%%%%%%%%%%

\begin{abstract}
    This paper describes an automatic method for \textit{adjusting} or \textit{tuning} models of multi-agent motion. Simulating the motion of bird flocks, human crowds, vehicle traffic, and other multi-agent systems is a widely used technique. These simulations model the behavior of a single group member (bird, human, or vehicle). The group behaviors (flock, crowd, traffic) \textit{emerge} from interactions between group members. These models typically have many numerical control parameters. Even if each parameter is intuitive in isolation, their interaction can be complex and nonlinear. It is challenging to determine which parameters to adjust for the desired change in group behavior. Changing one aspect of group behavior often causes other aspects to change, leading to a tedious process of incremental changes. This work takes an \textit{inverse design} approach. The desired group behavior is measured with a user-defined objective(/fitness/loss) function and optimized with a genetic algorithm. The objective function used here for basic flocking rewards proper spacing with neighbors, flying near a desired speed, and avoiding obstacles. Interestingly, the vivid alignment seen in bird flocks appears to emerge from maintaining proper spacing between flockmates.
\end{abstract}

\noindent{\small\textbf{Keywords:} multi-agent, inverse design, optimization, evolutionary computation, genetic algorithm, multi-objective,  boids, flocks, herds, schools, crowds, traffic. \textbf{Data/Code available at:} \url{https://github.com/cwreynolds/evoflock}}
\blfootnote{\textcopyright  2026 Craig Reynolds. Published under a Creative Commons Attribution 4.0 International (CC BY 4.0) license.}

%%%%%%%%%%%%%%%%%%%%%%%%%%%%%%%%%%%%%%%%%%%%%%%%%%%%%%%%%%%%

\begin{figure*}
    \centering
    \includegraphics[width=0.9\textwidth]{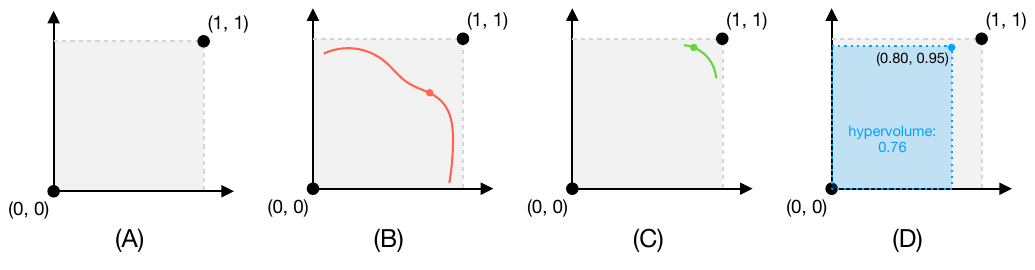}
    \caption{(A) Normalized multi-objective fitness space. For two objectives, a unit square. (B) The \textit{Pareto front} for two mutually conflicting objectives. (Like \textit{reliability} and \textit{affordability} in the hypothetical bridge design example.) The objectives can not all be fully satisfied at any point along the front. Visually: the ``distance'' from (1, 1) to the red curve. (C) Two ``mostly compatible'' objectives (which is to say: \textbf{not-}mutually-conflicting objectives) more typical of criteria for flock tuning. Here, the Pareto front passes near the global optimum. (D) Scalarization of a multiple-objective fitness value, using \textit{hypervolume}: the product of all objective scores. In this 2D example it is the blue rectangle's \textit{area}. Problems of type (C) can often be solved automatically with scalarized fitness. Problems of type (B) may require human judgment to select from among solutions along the Pareto front.}
    \label{fig:MOF_HV}
\end{figure*}

%%%%%%%%%%%%%%%%%%%%%%%%%%%%%%%%%%%%%%%%%%%%%%%%%%%%%%%%%%%%

\section{Introduction}
\label{sec:intro}

Simulation models of multi-agent motion are used in many fields including: animation, games, biology, robotic swarms, urban planning, training autonomous vehicles, and other applications. Since the 1980s, several models of group motion have been developed, including \textit{boids} and others (see \nameref{sec:related}). These allow creating simulations of flocks and other group motions, which most observers will recognize as some form of flocking. This paper will usually refer to bird flocks, with the assumption that other types of group motion (herds, schools, crowds, traffic, drone swarms) can be similarly modeled.

This project addresses the issue of \textit{adjusting} or \textit{tuning} multi-agent motion models toward a given behavioral goal. This is the \textit{inverse design} paradigm: formally state a goal, then use optimization to ``coax'' the model to match. For example, modifying an existing model of bird flocks to instead portray fish schools. Or starting from a model of flocking crows and changing it to flocking sparrows. Similarly, starting from a generic abstract flock model and fitting it to field observations of a particular species of real birds in nature. Or to take a plausibly realistic model of a natural bird flock, and change it, say for storytelling purposes, to convey a flock of birds that are happy, or angry. 

A boids-like simulation model typically has a collection of numeric parameters (``knobs'') that control its action. \textit{EvoFlock} is a software framework that automatically finds a set of near-optimal parameters for a simulation-based multi-agent system. The flock model used for these experiments is a predefined hand-written parametric ``black box'' model whose input is a \textit{parameter set} (consisting of 15 numbers, see Table \ref{table:flock-parameters}). The user provides a \textit{objective} (also known as \textit{fitness} or \textit{loss}) function. It takes a candidate parameter set, runs a simulation, and returns a \textit{score} reflecting how well the behavioral goals were met. The optimization process runs (for about two hours on a laptop) and produces a high quality parameter set, as measured by the user-supplied objective function (see e.g. Figure \ref{fig:boid_flock}).

For a user of this optimization framework, it is convenient that the flock model is treated as a black box. Almost no restrictions are imposed on a preexisting model by this optimization framework: not on model design, source code availability, or programming language. (As can be an issue if optimization uses gradient descent via automatic differentiation \citep{baydin_automatic_2018}.)

In the past, flock models have been tuned manually, often a frustrating and time-consuming process. For a new model, adjusting parameters is required simply to create group motion that looks like plausible flocking. Any modification to a group motion model (e.g. birds{$\rightarrow$}fish) requires further adjustments to the parameters. Each such adjustment requires selecting which parameter(s) to change and how much to increase or decrease them. The main difficulty is that effects of control parameters may overlap and interact in non-linear ways. Changing one usually requires changing others to compensate. Potentially many parameters need to be changed, leading to a tedious series of edits and tests.

This paper is about automating that adjustment process using metrics of flock quality and an optimization process. The metrics use multiple objectives. The nature of those objectives and how to add new ones will be described in sections \nameref{sec:FlockingObjective} and \nameref{sec:add_objective}. The optimization process used in these experiments is a genetic algorithm, described in section \nameref{subsec:Optimization_with_GA}.

A summary of (``hyper'')parameters for this framework is given in Table \ref{table:HyperParameters}. See the \href{https://github.com/cwreynolds/evoflock}{c++ code} for this project. \href{https://youtu.be/XkwJXPJwxIc}{Video 1} shows flock simulations based on this approach.

%%%%%%%%%%%%%%%%%%%%%%%%%%%%%%%%%%%%%%%%%%%%%%%%%%%%%%%%%%%%

\begin{figure*}
    \centering
    \includegraphics[width=0.7\textwidth]{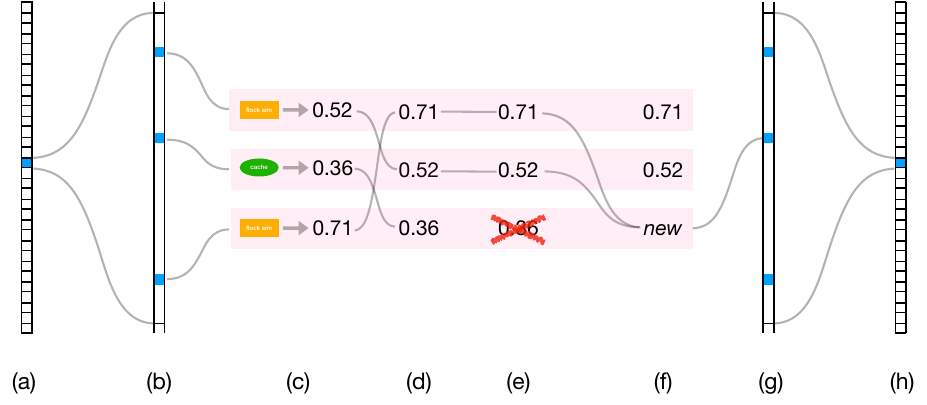}
    \caption{One update step of the (SSGA) evolutionary optimization process, using a 3-way tournament: (a) uniformly select one of the \textit{sub-populations}, (b) from that sub-population, uniformly select three \textit{individuals}, each holding a flock parameter set, (c) get fitness for each, either by running flock simulations (orange), or using the fitness cached from previous tournaments (green), (d) sort the three by fitness, (e) the worst (least fit) individual is deleted, (f) the best two individuals mate (\textit{crossover}) to create a new individual, (g,h) that new individual replaces the worst of the three back in the population. Removing the worst individuals, rather than promoting the best, is a form of \textit{negative selection} which tends to preserve diversity in the population.}
    \label{fig:evo_update}
\end{figure*}

%%%%%%%%%%%%%%%%%%%%%%%%%%%%%%%%%%%%%%%%%%%%%%%%%%%%%%%%%%%%

\section{Related Work}
\label{sec:related}

Since the 1980s, many simulation-based models of bird flocks have been proposed. They have broadly similar intent, with many differences in their modeling philosophy and computational technique. These models include \textit{boids} and many others:
\citet{aoki_simulation_1982}, 
\citet{akira_okubo_dynamical_1986}, 
\citet{reynolds_flocks_1987},
\citet{heppner_stochastic_1990}, 
\citet{tu_artificial_1994},
\citet{vicsek_novel_1995},
\citet{toner_flocks_1998}, 
\citet{couzin_collective_2002},
\citet{bajec_simulating_2005},
\citet{cucker_emergent_2007},
\citet{moskon_fuzzy_2007},
\citet{cavagna_seventh_2008},
\citet{bajec_organized_2009},
\citet{bhattacharya_collective_2010},
\citet{vasarhelyi_optimized_2018},
and
\citet{hoetzlein_flock2_2024}.

``Tuning'' the many parameters of these flock models can be difficult. EvoFlock seeks to help. It is a robust gradient-free evolutionary optimizer for parameter tuning to create (or improve or re-target) multi-agent motion models. EvoFlock handles large numbers of fast moving agents with potentially conflicting goals. It optimizes parameters to meet multiple procedural objectives, allowing for complex inverse design tasks. 

An early, but not very successful, attempt at evolutionary inverse design for flock simulations \citep{reynolds_evolved_1993} used \textit{genetic programming}. (GP is a variation of the \textit{genetic algorithm} (GA).) \citet{funes_shaping_2004} used interactive GA to find individual behaviors that produce ``interesting'' group behaviors, referring to this approach as an \textit{inverse problem}. \citet{stonedahl_finding_2011} use a GA to search parameter spaces using an objective function running flock simulations based on NetLogo \citep{tisue_netlogo_2004}. Similarly, \citet{demsar_evolution_2017} attempted ``to evolve collective behavior from scratch.'' \citet{vasarhelyi_optimized_2018} used a GA to tune flocking models suitable for the flight dynamics of a swarm of real drones. \citet{stolfi_escorting_2025} use evolutionary optimization to design robotic swarms for ``escorting'' tasks.

Other than those evolutionary-based solutions, most previous work on automatic tuning of flock/multi-agent models has used \textit{reinforcement learning} (RL) \citep{sutton_reinforcement_1998} as the optimization technique. For example, see recent excellent work on making high-quality \textit{driver agents} for a multi-agent traffic simulation \citep{cornelisse_building_2025} and \citep{cusumano-towner_robust_2025}. They use \textit{self-play} where agents learn (``bootstrap'') from interacting with each other. EvoFlock also has an aspect of self-play as flockmates learn to synchronize and cooperate with each other. A recently posted preprint uses reinforcement learning with an objective function \textit{very} similar to the one used here \citep{brambati_learning_2025}. See also research on drone-based communication networks \citet{sun_integrated_2025} that seek to maintain good spatial coverage while minimizing energy usage.

Close in spirit to EvoFlock is recent work on automatically tuning 2d herd models from overhead video \citep{gong_herds_2025}. From drone video of (say) sheep flocks, the system provides a dense simulated flock that reacts to simulated terrain quite like real sheep flocks would.

A recent example of controlling the dynamics of multi-agent motion using physically-based deep neural networks can be found in \citet{kim_commanding_2025}.

Notable, if somewhat tangential: \citet{jaderberg_human-level_2019} uses a hybrid approach to learn human-level game-playing agents for multiplayer games. It uses population-based reinforcement learning, which combines aspects of RL, population-based GA, and the self-play mentioned above. Also a flock-like variant on particle swarm optimization: \citet{hereford_flockopt_2011}. Similar in theme to EvoFlock, but for non-flocking but still \textit{self-organizing particle systems}: \citet{parkar_evolving_2024}.

The general topic of inverse design is related to the larger field of morphogenetic engineering \citep{doursat_morphogenetic_2012}. 

An analog of this work is described in \citet{el_saliby_eventually_2024} when using artificial neural networks for continuous control tasks, where the weights of a fixed neural architecture are efficiently tuned (learned) using evolutionary optimization.

EvoFlock was developed as part of a larger ongoing collaboration investigating new approaches to inverse design of multi-agent motion. This related work has not yet been published.

\section{Description of the Technique}
\label{sec:Description}

\begin{figure}
    \centering
    \includegraphics[width=0.9\linewidth]{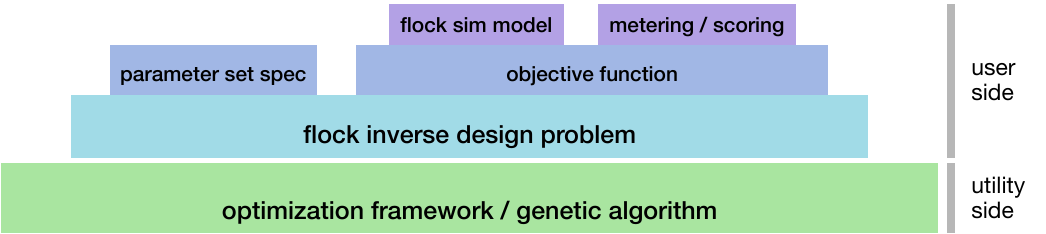}
    \caption{Overall structure of the flock optimization presented here. A generic optimization base (green) operates on a user's flock inverse design problem (blue). The user's description includes a parameter set specification and an objective function, the latter containing a flock simulator and score keeping utilities.}
    \label{fig:system_blocks}
\end{figure}

This section discusses the various components of the optimization process for adjusting parameters of a flock, or other kinds of multi-agent motion models. See Figure \ref{fig:system_blocks}.

\subsection{Optimization with Genetic Algorithms}
\label{subsec:Optimization_with_GA}

Optimization can be used to find a set of simulation parameters that best fit the behavioral goals, as given by an objective function. EvoFlock uses a \textit{genetic algorithm} (GA), a technique based loosely on concepts of biological evolution, as seen in the natural world. Evolution was first described by \citet{darwin_origin_1859} and \citet{wallace_tendency_1858}. Genetic algorithms were first described by \citet{holland_adaptation_1975}. A modern survey of the larger field of \textit{evolutionary computation} can be found in
\citet{de_jong_evolutionary_2016}.

A genetic algorithm maintains a \textit{population} of candidate solutions. Usually, the population contains a fixed number (tens to thousands) of candidate solutions, often called \textit{individuals}. Typically, each individual is a fixed-sized vector of numeric parameter values. The population may be further divided into \textit{demes} or \textit{islands}, representing semi-isolated breeding subpopulations. This is thought to promote diversity by pursuing different solutions in parallel. In an island model, there is usually some kind of \textit{migration} where individuals occasionally move from one island to another.

A GA is a population-based stochastic approach to optimization and discovery. Unlike random search, they are a random process guided by the frequency of useful partial solutions in the population, a model of biological \textit{gene frequency} in a species' \textit{genome}.

The ``genome'' of a GA changes over time under the influence of the objective function. For example, a model parameter might be changed by \textit{mutation}: adding a small signed zero-centered random value, to change it a bit from its previous value. Perhaps more importantly, individuals are modified by \textit{crossover} where two \textit{parent} individuals are recombined to create a new \textit{offspring}. In a rough model of biological crossover involving parental DNA, two parental parameter sets are merged into one, by taking each parameter at random from one parent or the other (called \textit{uniform crossover}).

Many variations on genetic algorithms have been developed over the years. Sometimes, the entire population of individuals is updated in parallel, as a \textit{generation}, and that process is repeated tens to hundreds of times. The approach used here, known as \textit{steady state genetic algorithms} (SSGA) updates individuals one at a time \citep{syswerda_study_1991}. If a generational GA is run for $G$ generations with a population of $P$ individuals, a corresponding SSGA runs for $G{\times}P$ steps. 

A preexisting GA framework called \textit{LazyPredator} is used for this project. See Figure \ref{fig:evo_update} for a diagram of the SSGA update cycle, and the ``tournament of three individuals'' based on \textit{negative selection} which promotes diversity in the population. LazyPredator's default 3-way tournament SSGA approach, and its hyperparameters, were used largely unchanged from previous unrelated projects.

%%%%%%%%%%%%%%%%%%%%%%%%%%%%%%%%%%%%%%%%%%%%%%%%%%%%%%%%%%%%

\begin{figure}
    \centering
    \includegraphics[width=\linewidth]{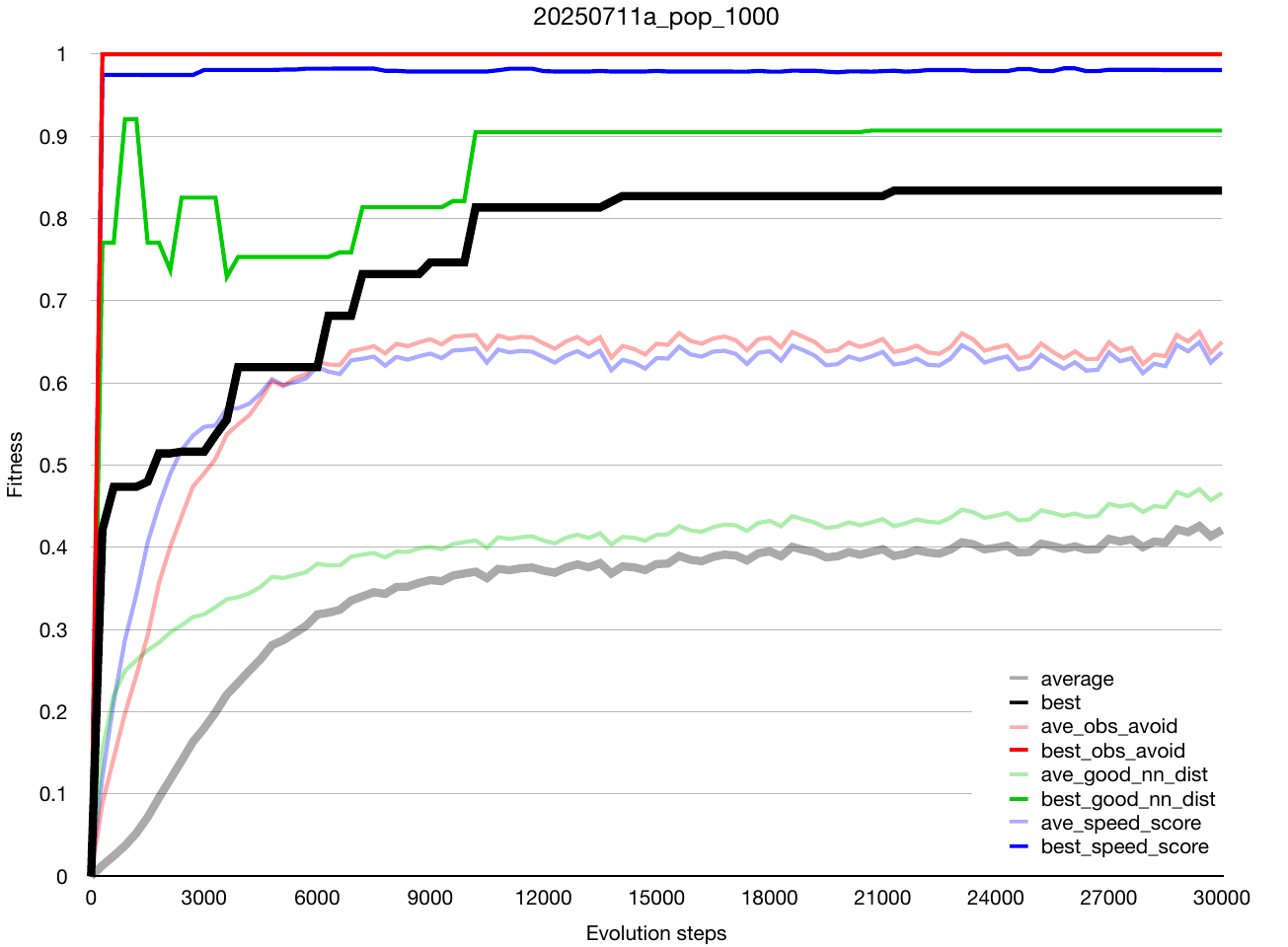}
    \caption{Plot of scalarized fitness (and its component metrics) over 30,000 steps of evolutionary time. The black bold trace (ending at 0.83) is the cumulative best population fitness. The bold gray trace (bottommost) is population average fitness. The colored traces correspond to best and average values (saturated and pastel) for the three component fitness objectives: obstacle avoidance (red), separation (green), and speed (blue). Data from run 20250711a.}
    \label{fig:fit_plot}
\end{figure}

%%%%%%%%%%%%%%%%%%%%%%%%%%%%%%%%%%%%%%%%%%%%%%%%%%%%%%%%%%%%

\subsection{Objective Function}
\label{subsec:ObjectiveFunction}

EvoFlock is based on optimizing a set of parameters for a flock (multi-agent motion) model. Optimization procedures are guided by an \textit{objective function}. In many optimization problems, the overall objective is a combination of several potentially contradictory goals. This is called \textit{multi-objective optimization} \citep{deb_multi-objective_2001}. 

To motivate this concept, consider building a bridge over a river. A key goal is that the bridge is reliable: that it will carry the required loads without collapsing. Another important goal is to minimize the cost of building the bridge. These criteria are directly opposed. A very strong bridge is reliable but costly. A very cheap bridge is unlikely to be reliable. This trade-off is often called \textit{Pareto optimality}. Tuples of such Pareto optimal values form the \textit{Pareto front}, see Figure \ref{fig:MOF_HV}(b).

\subsection{Multi-Objective Optimization}
\label{subsec:Multi-Objective}

Genetic algorithms were first used for problems with a single scalar fitness value. These have the advantage of having a \textit{total order}. However, like the bridge building example above, many problems inherently have \textit{multiple objectives} (see Figure \ref{fig:MOF_HV}(B)). For contradictory goals (e.g. reliable vs. cheap) a human expert may need to select a best solution of those along the Pareto front. Luckily, flock optimization problems addressed here seem to be in the ``mostly compatible'' type of multiple objective problems (that is, not-mutually-exclusive objectives, see Figure \ref{fig:MOF_HV}(C)) and so can be solved automatically with \textit{scalarization}. A \textit{scalarizer} maps from multiple objective values to a single scaler value that somehow summarizes the multiple objective values.

A simple and commonly used scalarizer is called \textit{hypervolume}, which simply multiplies together the components of a multiple-objective value. The name hypervolume refers to a \textit{n}-dimensional box whose edges correspond to the components of a multiple objective value (see Figure \ref{fig:MOF_HV}(D)). Hypervolume has the property that its value changes in proportion to changes in each component.

So, for example, the hypervolume goes up (or down) as any single component goes up (or down). This holds as long as none of the components is zero, which ``hides'' contributions of other components. To avoid this, the hypervolume variant used here remaps the fitness range from [0, 1] to [$\epsilon$, 1] before taking the product. ($\epsilon$ is 0.01 for the ``normalized fitness'' values used here.)

\section{Parameter set for a flocking model}
\label{sec:parameter_set}

The focus of EvoFlock is on \textit{inverse design}, the optimization of parameter sets to achieve group behavior under the direction of an objective function. As such, details of a particular flocking model are largely off-topic. However, just to ground the discussion for readers unfamiliar with flocking models, a quick overview is given here.

Most flock models are written from the perspective of an individual boid (agent) at a single time step. The boid identifies its neighbors. (The model used in this work uses \textit{topological distance} to find the seven nearest neighbors \citep{cavagna_seventh_2008}.) The boid reacts to its neighbors by computing \textit{steering forces} \citep{reynolds_steering_1999} related to their distance and angle. It may also need to steer to avoid obstacles. These component steering forces are combined (often by 3D vector addition) then applied to the boid's physics model (often a simple \textit{self-propelled particle} model).

%%%%%%%%%%%%%%%%%%%%%%%%%%%%%%%%%%%%%%%%%%%%%%%%%%%%%%%%%%%%

\begin{figure}[b!]
    \centering
    \includegraphics[width=0.9\linewidth]{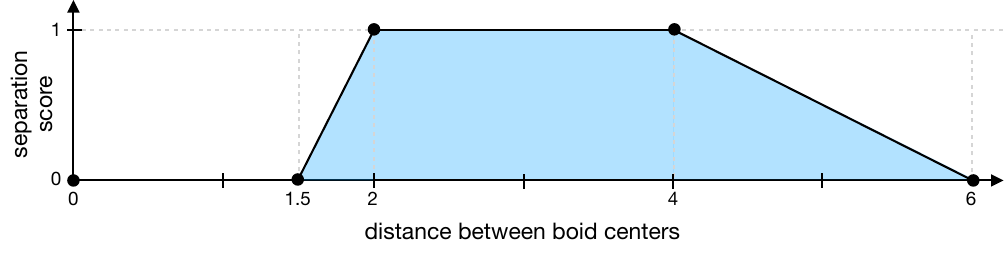}
    \caption{Score for desired \textit{separation} distance to a boid's nearest neighbor. This function is highest when the centers of two boids are between 2 and 4 body diameters apart. The default boid body diameter is 1. The \textit{separation} score for an entire flock simulation is the average of this function, over all \textit{boid-steps}. (That is, for 200 boids on 500 simulation steps, so 100,000 boid steps.)}
    \label{fig:SeparationScore}

    \centering
    \includegraphics[width=0.9\linewidth]{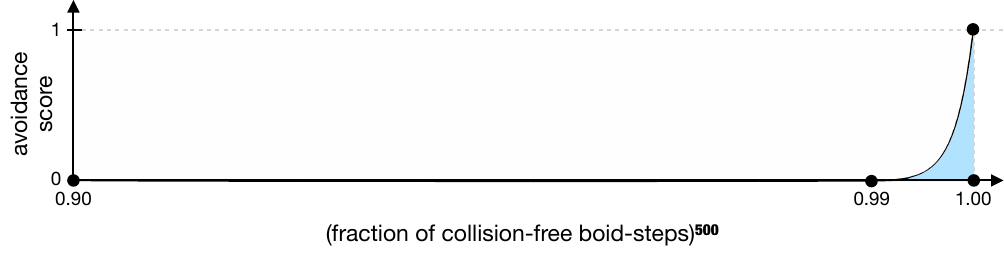}
    \caption{Score for obstacle avoidance. It is the fraction of collision-free boid-steps (on [0, 1]) raised to a high power (500). This function is nearly zero until the number of collision-free steps is about 99\%.}
    \label{fig:avoid_score}

    \centering
    \includegraphics[width=0.9\linewidth]{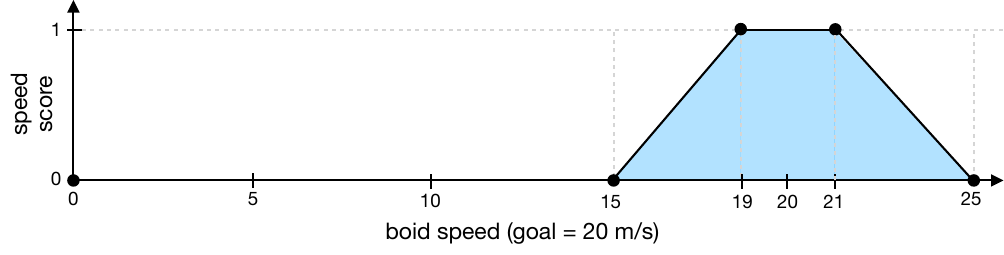}
    \caption{Score for desired boid \textit{speed}, here $\sim$20 meters per second. This function is highest when speed is between 19 and 21 m/s and falls off to zero outside that range. The \textit{speed} score for an entire flock simulation is the average of this function over all boid-steps.}
    \label{fig:speed_score}

\end{figure}

%%%%%%%%%%%%%%%%%%%%%%%%%%%%%%%%%%%%%%%%%%%%%%%%%%%%%%%%%%%%

Each of these steering behaviors has parameters, such as weights, distances, and angles. (For example, the weight given to alignment behavior, or the minimum time before a predicted obstacle collision at which the boid should steer to avoid.) There are also parameters related to its physical model, such as the maximum force it can apply.

The collection of all these make up the flock's \textit{parameter set} which is adjusted by optimization according the the objective function. See Table \ref{table:flock-parameters} for details of the parameter set used in these experiments.

\section{An objective function for flocking}
\label{sec:FlockingObjective}

Having described the components of EvoFlock (see \nameref{sec:Description}), this section focuses on the specific objective(/fitness/loss) function used to optimize a multi-agent motion model for flocking, see Figure \ref{fig:fit_plot}

The objectives described below are based on measurements made during a flock simulation. While the simulation is run, ``bookkeeping'' code gathers statistics. (This happens while updating each individual boid step, after a flock update step, and at the end of a simulation). Those metrics are recorded and accessed by the evolutionary optimization framework after it runs a flock simulation.

\subsection{Objective for Separation}
\label{subsec:separation_objective}

A key observation is that the birds in a flock are grouped closely together but nearly always avoid colliding with flockmates. These goals (close together, no collisions) can be seen as a requirement that the distance between simulated boids fall within a given distance interval. Specifically, for each boid, we determine its nearest neighbor and measure the distance between their centers. A score is computed on the basis of that distance. See Figure \ref{fig:SeparationScore}. It is zero if the distance is too small (potential collision) or too large (insufficient flock density). Within that acceptable distance range, its value is one. Outside that desired range, piecewise linear ramps transition down to zero.

An equally iconic property of natural flocks is that birds are \textit{aligned}, flying on nearly parallel paths with their neighbors. An interesting result of this work is that this alignment appears to \textit{emerge} from the ``acceptable neighbor distance'' metric. It was \textbf{not} necessary to have an explicit optimization objective for alignment. The optimization for proper separation appears to \textbf{cause} the alignment.

\subsection{Objective for Obstacle Avoidance}
\label{subsec:avoidance_objective}

The example application in this work can be described as ``flocking in the presence of obstacles'' as shown in Figure \ref{fig:boid_flock}. For collision avoidance, the goal is not to merely \textit{reduce} the number of obstacle collisions, but to effectively eliminate them. These fitness tests (typically) run 200 boids for 500 simulation steps. The relevant statistic is the total number of \textit{avoided} boid-obstacle collisions over 100,000 boid-steps in a flock simulation.

To strongly emphasize the importance of collision avoidance, the normalized obstacle avoidance count (on [0, 1]) is exponentiated to the 500\textsuperscript{th} power. This pushes nearly the whole score's range down to almost zero, except for a small region near a perfect score of one. This effectively rejects flock parameter sets with more than a handful of collisions per flock simulation. An informal threshold in these experiments has been {$\leq$}10 collisions per 100,000 boid-steps. See Figure \ref{fig:avoid_score}.

At each time step, part of each boid's simulation is predicting future obstacle collisions and applying steering force to avoid them. The same code detects obstacle avoidance failures. In simulation, this is a matter of a boid having incorrectly passed through the surface of an obstacle (a zero-crossing of the obstacle's \textit{signed distance function}). A kinematic constraint is invoked to move the boid back to the correct side of the boundary. This failure is recorded in a per-simulation collision counter.

\subsection{Objective for Flight Speed}
\label{subsec:speed_objective}

Early in these experiments (following early boid models \citep{reynolds_flocks_1987}) each boid's flight speed was kinematically clipped to remain below a maximum speed (20 meters per second). Later, a third flock optimization objective was added to establish a target speed range. A new behavior was added to each boid to apply a signed force along its direction of motion to adjust its flight speed toward the desired range. The strength of this behavior became one of the flock parameters to be optimized in the inverse flock design process.

Flight speed \textit{can} be left to ``float'' during optimization. Unfortunately, there is an annoyingly large basin of attractions around not moving (speed=0). It seems likely that any given motion-based inverse design problem would imply a range of valid speeds. This suggests that the speed range should be given in the inverse design problem statement. For example, modeling vehicle traffic and pedestrian crowds would imply very different speed ranges. Figure \ref{fig:speed_score} shows a speed score function used in these experiments. It has a target range in the interval [19, 21] with a fixed support ramping down to zero outside that range.

%%%%%%%%%%%%%%%%%%%%%%%%%%%%%%%%%%%%%%%%%%%%%%%%%%%%%%%%%%%%

\begin{figure}[b]
    \centering
    \includegraphics[width=0.9\linewidth]{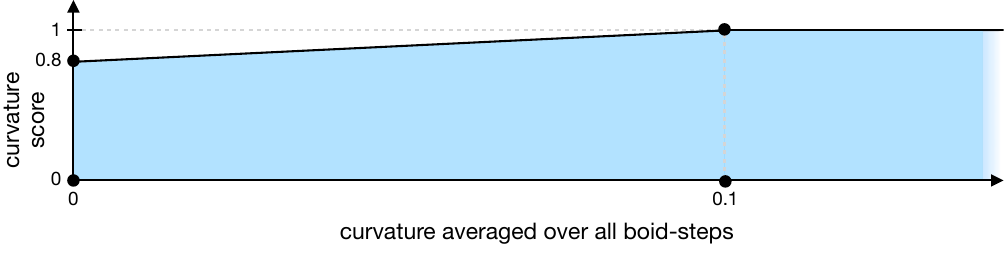}
    \caption{Score for boosting boid path curvature by slightly penalizing especially low curvature. Measured as average over all boid-steps of flock run. Score ramps from 0.8 to 1.0 for curvature on [0, 0.1] then clips at 1 for higher curvature.}
    \label{fig:curve_score}
\end{figure}

%%%%%%%%%%%%%%%%%%%%%%%%%%%%%%%%%%%%%%%%%%%%%%%%%%%%%%%%%%%%

%%%%%%%%%%%%%%%%%%%%%%%%%%%%%%%%%%%%%%%%%%%%%%%%%%%%%%%%%%%%

\begin{table}[t]
\centering
\begin{tabular}{ | l | c | c | }
    \hline
    \textbf{Description} & \textbf{Range} & \textbf{Units} \\
    \hline
    max force & [0, 100] & newtons \\
    % min speed & [20, 20] & \textit{unused} \\
    % speed & [20, 20] & \textit{unused} \\
    % max speed & [20, 20] & \textit{unused} \\
    forward weight & [0, 100] & \textit{none} \\
    separate weight & [0, 100] & \textit{none} \\
    align weight & [0, 100] & \textit{none} \\
    cohere weight & [0, 100] & \textit{none} \\
    predictive avoid weight & [0, 100] & \textit{none} \\
    static avoid weight & [0, 100] & \textit{none} \\
    max dist separate & [0, 100] & meters \\
    max dist align & [0, 100] & meters \\
    max dist cohere & [0, 100] & meters \\
    angle separate & [-1, +1] & cos(a) \\
    angle align & [-1, +1] & cos(a) \\
    angle cohere & [-1, +1] & cos(a) \\
    fly away max dist & [0, 100] & meters \\
    min time to collide & [0, 10] & seconds \\
    \hline
\end{tabular}
\caption{Flock model parameters subject to optimization: a collection of scalar values, each with an associated range. They are initially randomized then optimized over time according to the objective function. These values are held in a c++ class called \code{FlockParameters} along with ``constant'' parameters (flock population, time step, etc.) which are not subject to the optimization process.}
\label{table:flock-parameters}
\end{table}

%%%%%%%%%%%%%%%%%%%%%%%%%%%%%%%%%%%%%%%%%%%%%%%%%%%%%%%%%%%%

\section{Adding an Objective}
\label{sec:add_objective}

EvoFlock aims to simplify both creating and modifying flock models. This section describes such a modification, adding a new fitness objective to the ``flocking with obstacles'' behavior described in the previous section (three objectives: separation in range, speed in range, and avoid obstacles). The flocking motion that results is competent and smooth. Let us assume for some application-specific reason, the quality of motion is judged to be ``too smooth.''

This can be seen as a tendency of the previous three objectives to minimize the boid's \textit{path curvature}. Perhaps intentionally boosting the curvature would counteract that ``too smooth'' quality of motion? (For example, changing from ``boring'' to a more ``playful'' quality of motion?)

The \code{Boid} class used by EvoFlock already provided an accessor for its instantaneous path curvature. A candidate ``flock curvature'' score could be simply averaging that over all boid-steps. A test run showed that the instantaneous curvature of a boid's path is approximately on [0, 2]. The average curvature over a run is on [0, 1] at the beginning, but about [0, 0.04] at the end. This supports the notion that basic flocking minimizes curvature.

An objective metric for boosting curvature might map the average curvature of a run to a normalized fitness score. Prototyping this code change took about an hour of programmer effort, followed by a few two-hour optimization runs to find effective hyperparameters: a linear mapping (with clipping) from a curvature range of [0, 0.1] to a score range of [0.8, 1], see Figure \ref{fig:curve_score}. This is effectively a fitness penalty of 20\% for zero path curvature, ramping up to no penalty for ``quite curvy.'' 

A sample of the resulting flock motion can be seen in \href{https://www.youtube.com/watch?v=p8RE0ApsPuw}{Video 2}, a simulation with 1000 boids, using flock parameters evolved with this new curvature-boosting objective. (Compare with the non-curvature-boosting \href{https://youtu.be/XkwJXPJwxIc}{Video 1} mentioned in the \nameref{sec:intro}.)

%%%%%%%%%%%%%%%%%%%%%%%%%%%%%%%%%%%%%%%%%%%%%%%%%%%%%%%%%%%%

\begin{table}
\centering
\begin{tabular}{ | l | c | c | }
    \hline
    % \multicolumn{3}{|c|}{Project Parameters} \\
    \textbf{Description} & \textbf{Value} & \textbf{Notes} \\
    \hline
    boids per flock & 200 &  \\
    flock simulation steps & 500 & $\sim$16.6 sec \\
    % simulation time step & 1/30 seconds & \\
    % simulation time step & $\frac{1}{30}$ seconds & \\
    simulation time step & \sfrac{1}{30} seconds & \\
    \hline
    boid body diameter & 1 meter & \\
    boid body mass & 1 kilogram & \\
    boid max force & < 100 newtons & \\
    boid speed range & 19 to 21 m/s & \\
    obs avoid exponent & 500 & \\
    k nea\textbf{}rest neighbors & 7 & cf. StarFlag \\
    % k nearest neighbors & 7 & \citet{cavagna_seventh_2008} \\
    % k nearest neighbors & 7 & Cavagna (2008) \\
    \hline
    evolution population & 300 & individuals \\
    evolution steps & 30,000 &  SSGA steps \\
    equivalent generations & 100 & \\
    hypervolume epsilon & 0.01 & $\epsilon$ \\
    migrations per step & 0.05 &  \\
    \hline
    run time: flock sim & $\sim$0.25 seconds & $200{\times}500$ \\
    run time: evolution & $\sim$2 hours & \\
    \hline
\end{tabular}
\caption{Hyperparameters for the evolutionary inverse design framework. Run times are on a 2021 Apple Macbook Pro M1 Max. ``StarFlag'' is \citet{cavagna_seventh_2008}}
\label{table:HyperParameters}
\end{table}

%%%%%%%%%%%%%%%%%%%%%%%%%%%%%%%%%%%%%%%%%%%%%%%%%%%%%%%%%%%%

\section{Conclusions}
\label{sec:Conclusions}

This work shows that evolutionary optimization, via a genetic algorithm, can successfully find a set of parameters for a ``black box'' boid flocking model. This suggests that other similar multi-agent motion models --- for example, multi-robot terrain search --- might also be solved using a genetic algorithm in a similar way, using an objective with some similarity to those discussed here.

One advantage of using evolutionary optimization for this type of inverse design problem is that the multi-agent simulation can be treated as a ``black box.'' As such, there are few constraints on how the simulation is implemented. In contrast, using \textit{stochastic gradient descent} (SGD) \citep{robbins_stochastic_1951} requires that the flock model be differentiable, either analytically or via \textit{automatic differentiation} \citep{baydin_automatic_2018}. Conveniently, evolutionary optimization only requires only a flock simulation's scalar fitness value, no gradients are needed.

A surprising finding of this work is that the \textit{alignment} of birds in a flock seems to \textit{emerge} from maintaining proper spacing between flockmates. In retrospect this seems plausible: a group of fast moving animals, with a desire to remain close-but-not-too-close to each other, while dodging around obstacles, must travel on parallel curved paths. However, before these experiments, it was not clear whether an objective function specifically to require alignment would be needed.

A note about GA population size versus the number of SSGA updates (see Figure \ref{fig:evo_update}). Some experiments were made increasing the evolutionary population from 500 to 1000, while holding the total number of SSGA updates constant at 30,000. The hope was that more \textit{individuals} (more flock parameter sets) might allow a better solution to be found by using a wider coverage of the parameter space. The opposite seemed to be true. Larger populations led to worse results (often summarized as the best scalar fitness at end of run). In contrast, reducing the population from 500 to 300 produced somewhat better results, but 150 was worse. Finally, for a population size of 300, more steps seemed to help: running 30,000 updates produced fitness around 0.82, while running 60,000 updates produced fitness around 0.88. Apparently, having more updates per individual produces better results.

%%%%%%%%%%%%%%%%%%%%%%%%%%%%%%%%%%%%%%%%%%%%%%%%%%%%%%%%%%%%

\begin{figure}
    \centering
    \includegraphics[width=0.9\linewidth]{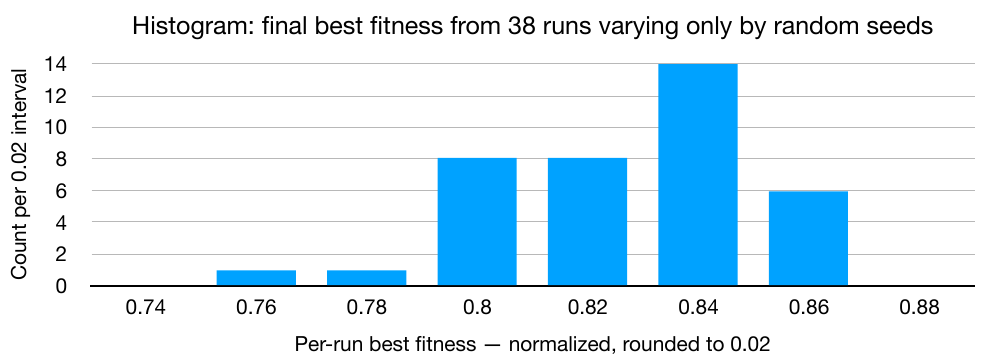}
    \caption{A histogram of final best fitness from 38 identical runs, varying only by random seeds}
    \label{fig:seed_variance}
\end{figure}

%%%%%%%%%%%%%%%%%%%%%%%%%%%%%%%%%%%%%%%%%%%%%%%%%%%%%%%%%%%%

\section{Limitations}
\label{sec:limitations}

Because EvoFlock is a stochastic algorithm, the quality of the results varies from run to run. (For the three-objective version, typical end-of-run best fitness is around 0.83, but can be as low as 0.76. See Figure \ref{fig:seed_variance}.) This procedure is not deterministic, both because the random number generator is seeded from wall clock time for each run and because the objective function is multithreaded and so is inherently nondeterministic.

The current objective function runs \textbf{four} flock simulations for each parameter set. These are run in parallel threads on separate CPU cores, so there is little time penalty. However, if not for this 4{$\times$} load, the cores would be free for other types of parallelism within one flock simulation. The four simulations are an attempt to reduce fitness variance of the randomized simulations, to avoid ``lucky'' high scores. (Of the four fitness values, the \textit{lowest} (worst) is returned from the objective function as a conservative estimate.) Recent work \citep{antipov_evolutionary_2025} has shown that this extra computation may be misguided, and that the overall quality of results do not suffer from using noisy raw fitness values.

\section{Future Work}
\label{sec:future}

The examples in this work were based on tightly enclosed spaces cluttered with obstacles, as in Figure \ref{fig:boid_flock}. These environments mean boids are always in ``emergency mode'' trying to avoid collisions with obstacles or with their flockmates. What happens when these constraints are relaxed? Other researchers have looked at ``open space flocking'' \citep{hoetzlein_flock2_2024} and \citep{brambati_learning_2025}. A parameter set found by EvoFlock, for crowded environments, has disappointing behavior when applied to open space. Small ``subflocks'' head off in different directions without further interaction. This is especially true for the 3 objective version (separation, speed, obstacle), and to a lesser extent, the 4 objective version (with curvature). Digging deeper into open space flocking, such as ``murmurations,'' seems a worthwhile direction for future work. A work-in-progress test using EvoFlock for open space murmuration can be seen in \href{https://www.youtube.com/watch?v=n-IKdgAOzSA}{Video 3}.

Similarly, multi-agent models with non-uniform agents \citep{montanari_optimal_2025} and non-reciprocal behaviors \citep{choi_flocking_2025}, \citep{weis_generalized_2025} are promising future research topics. Potential applications could include predator-prey models or automobile-bike-pedestrian traffic models.

These evolutionary techniques may help tune parameters for modeling various types of \textit{stigmergic} behavior, as when social insects collaborate, via signals in the environment, to build communal structures: ant colonies, termite mounds, bee honeycomb \citep{camazine_self-organization_2003}.

The formulation of speed score (Figure \ref{fig:speed_score}) and separation score (Figure \ref{fig:SeparationScore}) include ``ramps'' outside the target range. Are these necessary or even helpful? The robust sampling provided by 100,000 scores per run may tend to ``smooth out'' those scores without the extra complexity of choosing slopes for ramps.

EvoFlock finds good results using the genetic algorithm \citep{holland_adaptation_1975}. It would be interesting to try a version using genetic programming \citep{koza_genetic_1992}. This would eliminate the need to start by hand coding a ``black box flocking model'' and could perhaps evolve it from scratch.

An anonymous reviewer suggested that the steady-state genetic algorithm used here may not perform as well as continuous space evolution strategies \citep{hansen_coco_2021} such as ``Covariance Matrix Adaptation Evolution Strategy''. However, the SSGA used here seemed to perform well in this application, for both quality of results and speed of optimization. As mentioned just above, a future goal of this work is to skip the stage of writing black-box models by hand by using genetic programming. The optimizer framework used here, LazyPredetor, is based on GP, performing GA via a source level transformation.

\section{Acknowledgements}
\label{sec:ack}

This work is part of a larger project. Many thanks for conversations and camaraderie to my collaborators: Gilbert Bernstein, Matthew Shang, Jennifer Luo, Ryan Zambrotta, and Tzu-Mao Li. Thanks to Ronin Barzel for a careful review of an early draft of this paper. I also want to thank my long time (30 year?) friend and ALife colleague, Inman Harvey, whose deep insights taught me so much about evolutionary computation and scholarship. \textit{[Note: I wrote that in an early draft in 2025. Now in June 2026 as I prepare this final draft, comes the sad news that Inman has passed away. May his memory be a blessing.]} To Andy Kopra for discussions that led to my 1993 paper. Thanks to John Holland for genetic algorithms and to John Koza for introducing me to GP and GA. Thanks to the anonymous reviewers for helpful suggestions. Finally, to my family for their loving support.

\bibliographystyle{apalike}
\bibliography{EvoFlock.bib}

\end{document}